\documentclass[conference]{IEEEtran}
\usepackage{cite}
\usepackage{amsmath,amssymb,amsfonts}
\usepackage{algorithmic}
\usepackage{graphicx}
\usepackage{textcomp}
\usepackage{xcolor}
\usepackage{comment}
\usepackage{multirow} 
\usepackage{booktabs}
\usepackage{makecell}
\usepackage{capt-of}
\usepackage{cuted}
\begin{document}

\title{HyperSim: A Holistic Sim-To-Real Framework For Robust Robotic Manipulation}
\author{
\IEEEauthorblockN{Junyi Dong$^{1, *,\dag}$, Haotian Luo$^{1,*}$, Ziwei Xu$^{1}$, Shengwei Bian$^{1}$, Heng Zhang$^{1}$, Sitong Mao$^{1}$, Jingyi Guo$^{1}$, \\
Yang Xu$^{1}$, Wenhao Chen$^{1}$, Qiuyu Feng$^{1}$, Yao Mu$^{2}$, Ping Luo$^{3}$, Shunbo Zhou$^{1}$, and Xiaodong Wu$^{1}$}\\
\IEEEauthorblockA{$^{1}$ CloudRobo Lab, Huawei Cloud Computing Technologies Co.,Ltd.,  \\ $^{2}$ Shanghai Jiao Tong University 
$^{3}$ The University of Hong Kong} 
{$^{*}$ Equal contribution  $^{\dag}$ Corresponding authors}\\
}


\twocolumn[{
  \begin{@twocolumnfalse}
    \maketitle
    \centering
    \includegraphics[width=0.93 \textwidth]{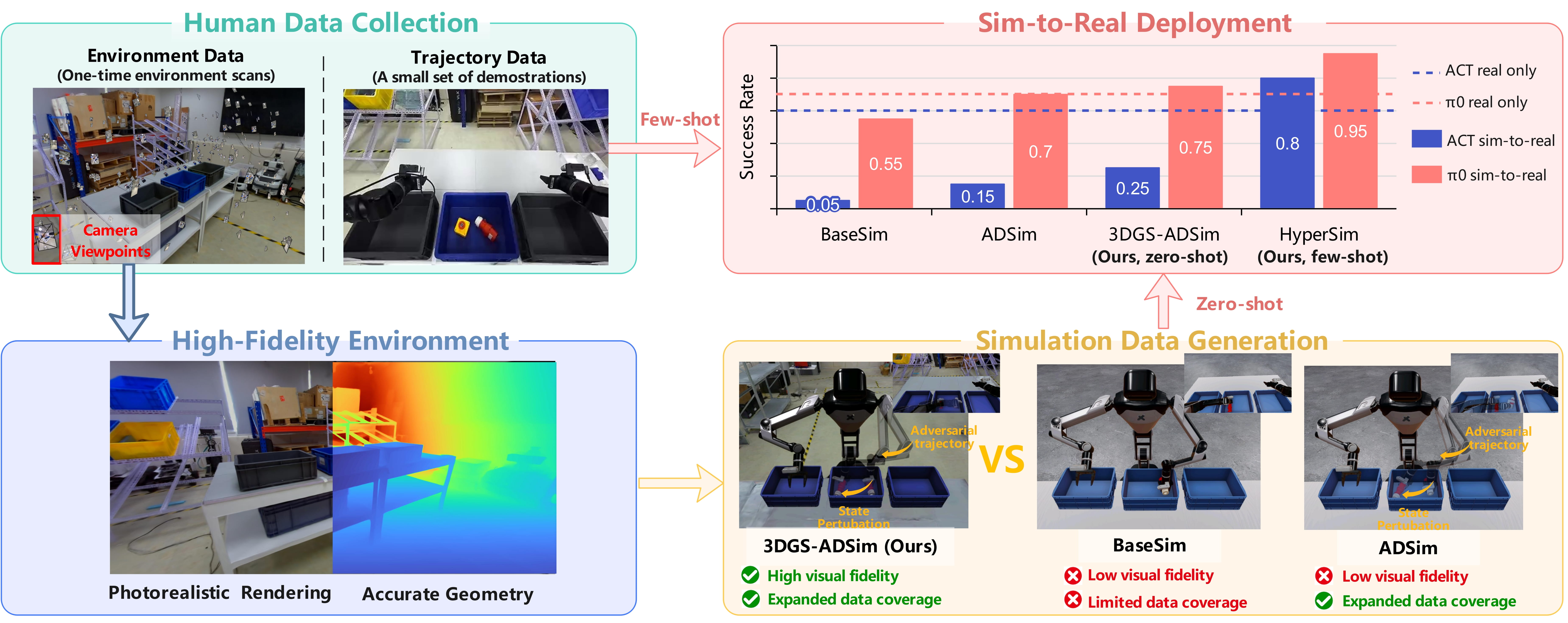}
    \captionof{figure}{Requiring minimal human-collected data (a one-time environment scan and a few dozen demonstrations), our method reconstructs photorealistic and geometrically accurate environments to generate adversarial trajectories for policy training. Evaluated using both ACT and $\pi_0$ policies, our approach consistently outperforms the ablated baselines under both zero-shot and few-shot deployments.}
    \label{fig:DataToPolicy}
    \vspace{10pt} 
  \end{@twocolumnfalse}
  }]

\begin{abstract}
Scaling data volume and diversity is critical for generalizing embodied intelligence. While synthetic data generation offers a scalable alternative to expensive physical data acquisition, transferring robotic manipulation policies from simulation to the real world (sim-to-real) remains a formidable challenge due to the domain gap. 
This paper presents HyperSim, a holistic framework spanning from synthetic data generation to policy training and seamless real-world deployment.  To systematically bridge the sim-to-real gap, HyperSim is realized through three core pillars: high-fidelity environment synthesis, adversarial trajectory generation, and sim-and-real co-training. 
Collectively, these modules address domain discrepancies by enhancing visual fidelity, expanding data coverage, and enforcing domain-invariant representations. We rigorously validate HyperSim through a large-scale empirical study involving 400 real-world task executions across two representative manipulation models. Assessed across three fine-grained metrics, our complete pipeline achieves remarkable sim-to-real success rates of 80\% and 95\% with ACT and $\pi_0$, respectively. Furthermore, policies trained on our adversarial trajectories exhibit significantly enhanced robustness against dynamic uncertainties, achieving a 35\% higher completion rate under physical perturbations.

\end{abstract}

\section{Introduction}
Driven by recent breakthroughs in multimodal large models, the field of robotics is undergoing a paradigm shift from modular skill learning to data-driven, end-to-end policy learning \cite{KimOpenVla24, LiuRDT24}. However, unlike foundation models in vision and language that can directly leverage vast internet-scale data, embodied agents require paired action-observation sequences that encompass detailed physical interactions. Acquiring such high-quality robotic datasets necessitates substantial investments in hardware infrastructure and human labor \cite{NasirianyRoboCasa24, WuRoboMind25, FanTwinAligner25}. Consequently, with the exponential growth in model parameters and data hunger of emerging robot foundation models \cite{Pi05_25, GeneralistGen0_25}, relying exclusively on real-world data collection has become increasingly impractical.

Synthetic data offers a promising solution to circumvent this data bottleneck. While automated generation pipelines have made data acquisition in simulation highly efficient, existing methodologies \cite{WangRoboGen23, MandlekarMimicgen23, HaScalingUp23, NasirianyRoboCasa24, HuaGenSim25, ChenRoboTwin25, WuAscent25} suffer from three critical limitations.
First, simulation environments are often oversimplified, typically featuring a floating table against a void background, which fails to reflect the cluttered and unstructured nature of real-world environments.
Second, data diversity is constrained due to heuristic scene configurations  (e.g., fixing object orientations to artificially reduce task complexity)  and the predominance of homogeneous, successful-only trajectories. Together, these factors severely restrict the coverage of the state-action manifold.
Third, the persistent visual and dynamic mismatch between simulation and reality continues to degrade real-world performance.
Consequently, policies trained solely on simulation data often fail to generalize to physical environments, leaving effective sim-to-real transfer a  non-trivial challenge.

To overcome the aforementioned limitations, we introduce HyperSim, a holistic framework spanning from synthetic data generation to policy training and seamless real-world deployment. HyperSim features a two-layer design that strategically separates scalable data generation from advanced domain transfer. The base layer establishes a  procedural workflow: it synthesizes physically plausible and diverse scenes, and decouples complex tasks into distinct motion and interaction primitives for efficient trajectory generation.
Building upon this, the enhancement layer adopts a modular paradigm to systematically bridge the sim-to-real gap through three mechanisms.
First, to mitigate visual discrepancies, we employ geometry-aware Gaussian Splatting (GS), rendering environments with photorealism and geometric fidelity superior to traditional meshes.
Second, we inject state perturbations during trajectory synthesis to capture observations from diverse viewpoints and record recovery behaviors. This adversarial mechanism expands the state-action coverage for policy robustness.
Finally, a sim-and-real co-training strategy is applied to enforce learning of domain-invariant representations across paired observations and actions. 
We validate the proposed framework through extensive physical experiments on the Galaxy R1 platform. Assessed across a set of fine-grained metrics, HyperSim demonstrates robust sim-to-real transferability.

In summary, our main contributions are as follows.
\begin{itemize}
    \item  \textbf{We introduce HyperSim}, a comprehensive framework that systematically bridges the sim-to-real gap via high-fidelity scene construction, expanded state-action coverage, and cross-domain co-training.
    \item \textbf{We propose a hybrid scene synthesis strategy} alongside an \textbf{adversarial trajectory generation mechanism}. Together, they efficiently procedurally generate diverse, photorealistic scenes while injecting dynamic physical perturbations to expand the state-action manifold and enforce policy recovery.
    \item  \textbf{We demonstrate exceptional real-world transferability through systematic physical evaluations}. Our analysis reveals that policy performance is positively correlated with synthetic data fidelity and distribution coverage, and is further boosted by pre-trained foundation models and co-training. Overall, the complete HyperSim pipeline achieves an impressive zero-shot success rate of  75\% and few-shot success rate of 95\% with $\pi_0$ policies. In addition, our adversarial mechanism endows the policies with enhanced robustness, yielding a 35\% improvement under external physical disturbances.
\end{itemize}

\section{Related Work}

\subsection{Environment Construction For Robotics Simulation}
High-fidelity simulation environments are pivotal for mitigating the visual sim-to-real gap in embodied learning. Existing approaches face a fundamental trade-off between visual realism and physical interactivity, occupying two ends of a spectrum. At one end, procedural generation techniques \cite{ChenRoboTwin25, WangRoboGen23, TianInternDataA1_25}  
excel in scalability and are well-suited for synthesizing physically interactive scenes. However, they often lack photorealism and rely on randomization strategies that fail to capture the unstructured clutter of the real world. At the opposite end, neural rendering methods such as 3D Gaussian Splatting (3DGS) \cite{3dgs, 2dgs, pgsr} 
offer unparalleled photorealism but are hampered by ambiguous scale estimation and noisy geometric meshes, which are detrimental to contact-rich simulations \cite{robosplat}. In contrast, we propose a hybrid approach  that leverages geometry-aware 3DGS solely for complex background reconstruction while utilizing a constraint-aware generator for the foreground manipulation zone. This ensures both visual-physical alignment and the diversity of manipulation workspaces. 

\subsection{Scaling Trajectory Generation In Simulation}
Unlike computer vision and natural language processing, where models are typically trained on images or texts, embodied intelligence relies on paired action-observation sequences, commonly referred to as trajectories. As such, trajectory generation represents a unique and pivotal challenge fundamental to embodied AI research.
Prior efforts can be broadly categorized into demonstration augmentation
and piece-wise generation. 
The augmentation approaches exemplified by MimicGen and its successors \cite{JohnsSingleDemonstration21, DiSelfReplay22, MandlekarMimicgen23, NasirianyRoboCasa24} expand a few human demonstrations into thousands of trajectories by applying spatial transformation and interpolation to seed data.
However, such augmentation strategy is prone to generate non-smooth trajectories with discontinuous changes. 
Conversely, piece-wise generation like RoboTwin \cite{ChenRoboTwin25}, RoboGen\cite{WangRoboGen23} and InternData-A1 \cite{TianInternDataA1_25} 
decompose tasks into distinct motion and skill primitives. These primitives are then independently synthesized via motion planners, constrained optimization, or reinforcement learning leveraging privileged simulation states.
Our method falls into the second category with a novel design that injects perturbations into object states and induces recovery motions. The resulting adversarial trajectories enable robots to acquire richer observations that are crucial for policy generalization.

\subsection{Sim-To-Real Transfer}
The objective of sim-to-real transfer is to deploy robotic control policies trained in simulation to physical hardware by mitigating the inherent visual and dynamic discrepancies between the two domains.
To achieve this, existing literature predominantly relies on domain randomization \cite{ChenRoboTwin25} or system identification \cite{ChenRoboTwin25}. However, these methods often require  domain expertise and tedious trial-and-error to tune parameters such as randomization ranges or dynamics coefficients. 
Alternatively, co-training adopts a different philosophy by jointly optimizing robot policies over a mixture of simulated and real-world data \cite{WeiCotrainingDiffusionPolicy25, MaddukuriCoTrainingRecipe25, YangInvarianceCotraining25}, forcing the extraction and learning of domain-invariant representations.
Our key insight is that the brittleness of the robot policy in real world generalization arises from the combined factors such as mismatches in 
environment complexity and data distribution, as well as the visual and physics discrepancy. Therefore, we seek to address these issues synergistically in a holistic framework with systematic experiments and investigations, rather than individually handling them.

\section{The Proposed Approach} 
HyperSim is a holistic framework encompassing synthetic data generation, policy training, and real-world deployment (Fig. 2). To systematically close the sim-to-real gap, it integrates three functional modules: high-fidelity scene construction for visual alignment, adversarial trajectory synthesis for expanded state-action coverage, and co-training for domain-invariant representations. To seamlessly accommodate these capabilities, HyperSim adopts a flexible two-layer architecture. Specifically, a base layer drives the standard data pipeline, while a customizable enhancement layer flexibly integrates the aforementioned advanced sim-to-real transfer techniques. We elaborate on these three functional components below.

\begin{figure*}[t]
	\centering
	\includegraphics[width=0.92 \textwidth]{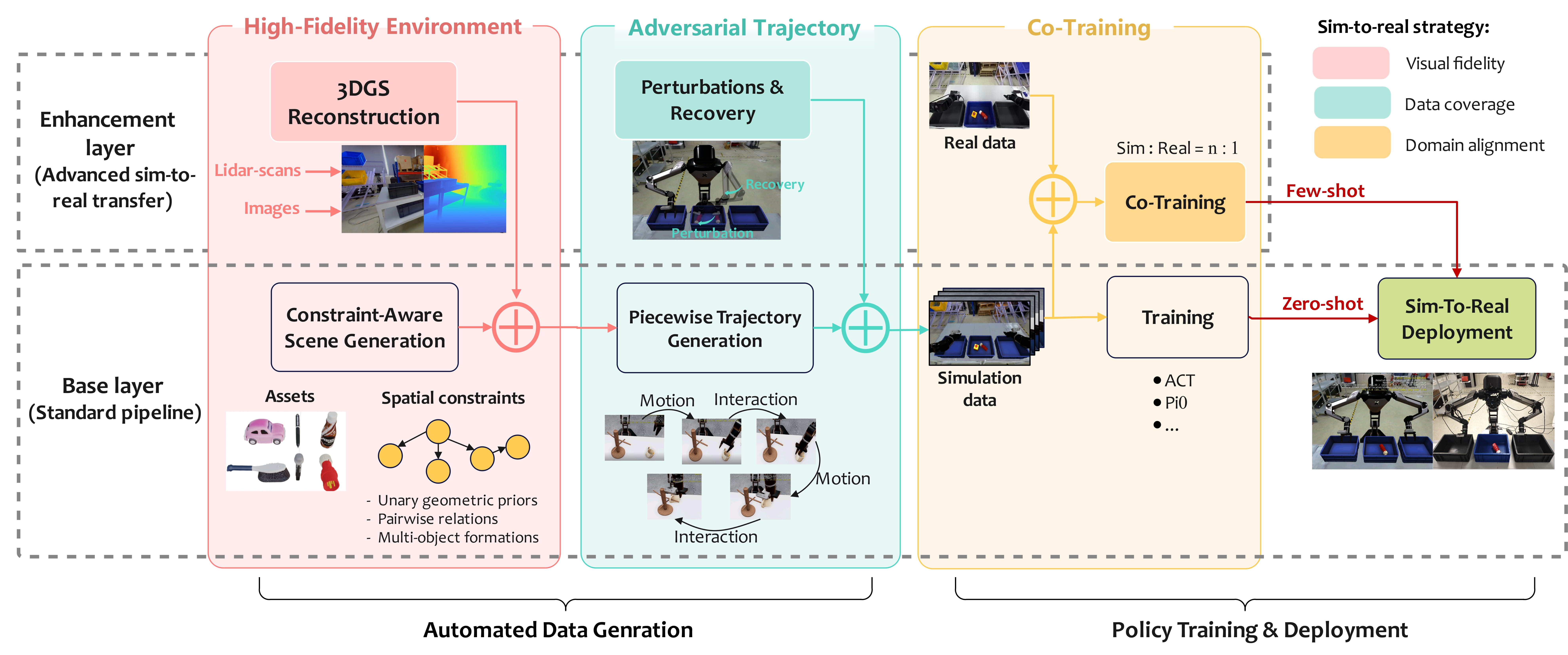}
	\caption{\textbf{Overview of the HyperSim framework}. HyperSim couples a standard data-to-policy pipeline with an enhancement layer to support physical deployment. By integrating high-fidelity environments, adversarial trajectories, and co-training, HyperSim systematically bridges the sim-to-real gap in visual fidelity, data coverage, and cross-domain feature representation}
	\label{fig:Framework}
\end{figure*}

\subsection{High-Fidelity Simulation Environment}
\label{sec:Environment}
Our approach introduces a hybrid synthesis strategy where the simulation environment explicitly consists of two components: a foreground manipulation area and a surrounding background.  This decoupling allows us to efficiently generate diverse, physically plausible workspaces in the foreground, which are seamlessly embedded within the photorealistic and geometrically accurate background.

\begin{figure}[thpb]
	\centering
	\includegraphics[width=0.38 \textwidth]{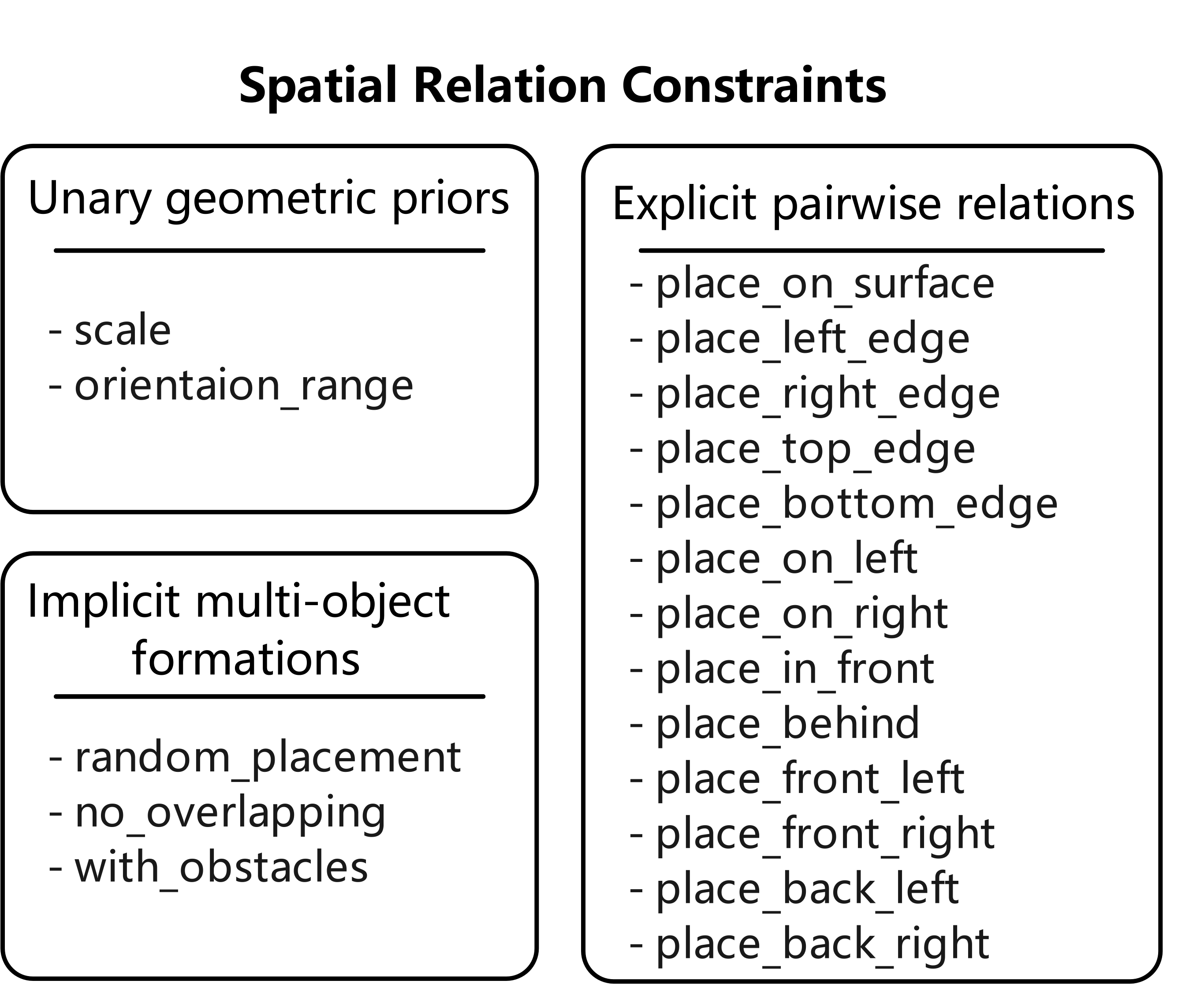}
	\caption{\textbf{Spatial relation constraints} used for foreground scene generation.}
	\label{fig:FigSpatialRelations}
\end{figure}

For the interactive foreground, we move
beyond randomized scene generation to enable precise spatial relation modeling. As illustrated in Fig.\ref{fig:FigSpatialRelations}, we design a library of 18 solvers which are categorized into three groups: unary geometric priors, explicit pairwise relations, and implicit multi-object formations.
The task-specific foreground requirements are translated into a subset of these spatial constraints to generate a valid layout. This synthesized layout is subsequently populated with high-fidelity 3D assets retrieved from established databases or generated by text-to-3D models to ensure physical interactivity. 
Detailed constraints explanations are deferred to the Appendix \ref{app:foreground}.

Complementary to the interactive foreground, our background reconstruction minimizes the visual sim-to-real gap. We collect synchronized multimodal data (RGB, LiDAR, and IMU) and leverage GPGS \cite{gpgs}—an state-of-art approach which exploits geometric priors derived from fused LiDAR scans, to reconstruct the environment into a set of Gaussian primitives. Incorporating geometric priors effectively alleviates the challenge of precise surface reconstruction for weakly textured or complex geometries, overcoming a key limitation of image-only approaches. Guided by the optimization scheme proposed in \cite{gpgs}, the Gaussian primitives are spatially aligned with the surfaces defined by fused LiDAR scans, enabling high-fidelity novel view synthesis of both color and depth maps.
The parameters of Gaussian primitives, including their position $\mathbf{p}_k$, covariance matrix $\Sigma$, opacity $o_k$ and spherical harmonics coefficients $\mathbf{c}_k$ which encapsulate view-dependent appearance, can be calculated by \cite{3dgs}:
\begin{equation}
    \mathcal{G}(\mathbf{p}) = exp (-\frac{1}{2}(\mathbf{p} - \mathbf{p}_k)^T \Sigma^{-1} (\mathbf{p} - \mathbf{p}_k))
\end{equation}
During volumetric rendering, each 3D Gaussian is projected to the image plane as $\mathcal{G}^{2D}$, then depth-sorted and alpha-blended into the final image expressed as pixel color values:
\begin{equation}
    \mathbf{c}(\mathbf{x}) = \sum_{k=1}^K \mathbf{c}_k o_k \mathcal{G}^{2D}(\mathbf{x}) \prod_{j=1}^{k-1} (1 - o_j\mathcal{G}^{2D}(\mathbf{x}))
    \label{eq:rendering}
\end{equation}

The pixel depth values can be rendered by sharing a similar strategy as above. Once the Gaussian representation is constructed, we leverage TSDF algorithm to fuse rendered color and depth maps and generate a colorized mesh structurally aligned with the Gaussian representation. Crucially, this hybrid Gaussian-mesh representation natively integrates with standard physics simulators: the 3D Gaussian splats facilitate photorealistic rendering, while the strictly aligned underlying meshes serve as a robust backend for precise collision and contact dynamics.

\subsection{Adversarial Trajectory}
\label{sec:AdversarialTrajectory}
To generate adversarial trajectories, we first decompose manipulation tasks into piecewise motion primitives. Subsequently, we introduce a controlled perturbation-and-recovery process during trajectory synthesis, which expands the state-action distribution and induces recovery behaviors.

\subsubsection{Piecewise Trajectory Generation}
\label{sec:Piecewise}

\begin{figure}[thpb]
	\centering
	\includegraphics[width=0.38 \textwidth]{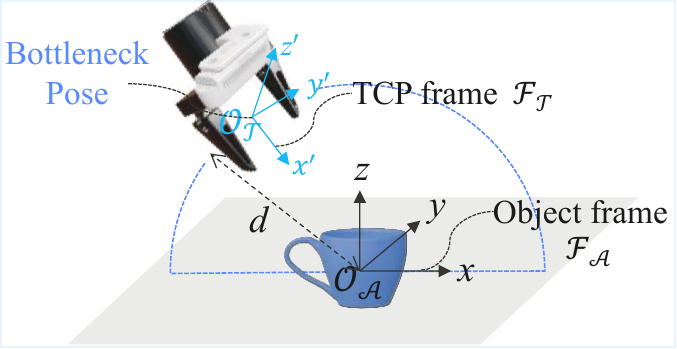}
	\caption{\textbf{ The bottleneck pose and local coordinate frames}. The  gripper TCP enters a small hemisphere of radius $d$ with respect to the target object.}
	\label{fig:FigBottleneckPose}
\end{figure}

Inspired by prior literature \cite{JohnsSingleDemonstration21,DiSelfReplay22, MandlekarMimicgen23}, we decompose manipulation tasks into a sequence of object-centric subtasks. 
Let $\mathcal{F}_{A}$ denote the object-centric Cartesian frame, with its origin $\mathcal{O}_{A}$ located at the target object's center of mass and its $xy$-plane parallel to the workspace surface. Assuming the use of a standard two-finger gripper, we define the tool frame $\mathcal{F}_{T}$ with its origin $\mathcal{O}_{T}$ at the Tool Center Point (TCP). By convention, the $x$-axis points toward the fingertips, the $y$-axis aligns with the gripper stroke connecting the two fingers, and the $z$-axis follows the right-hand rule, as shown in Fig.\ref{fig:FigBottleneckPose}. 

A typical manipulation subtask begins with a collision-free motion toward the target, which transitions into a transient state before physical contact occurs. Inspired by prior works \cite{JohnsSingleDemonstration21, DiSelfReplay22}, we formalize this transient state as the \textit{bottleneck pose}, defined as the specific configuration where the TCP enters a small hemisphere of radius $d$ centered at $\mathcal{O}_{A}$ ( Fig.\ref{fig:FigBottleneckPose}). 

Defining the bottleneck pose effectively partitions a manipulation subtask into two phases: an \textit{approaching primitive} spanning from the initial state to the bottleneck pose, and a subsequent \textit{interaction primitive} governing the object contact and manipulation.
Consequently, complete subtask trajectories can be synthesized in a \textit{piecewise} manner utilizing a combination of motion planners, inverse kinematics solvers, and gripper controllers \cite{ChenRoboTwin25, TianInternDataA1_25}. An illustrative example of these segmented primitives is provided in Fig.\ref{fig:TrajDecomposition}.

\subsubsection{Adversarial Perturbation And Recovery}
\label{sec:Adversarial}
\begin{figure}[thpb]
	\centering
	\includegraphics[width=0.48 \textwidth]{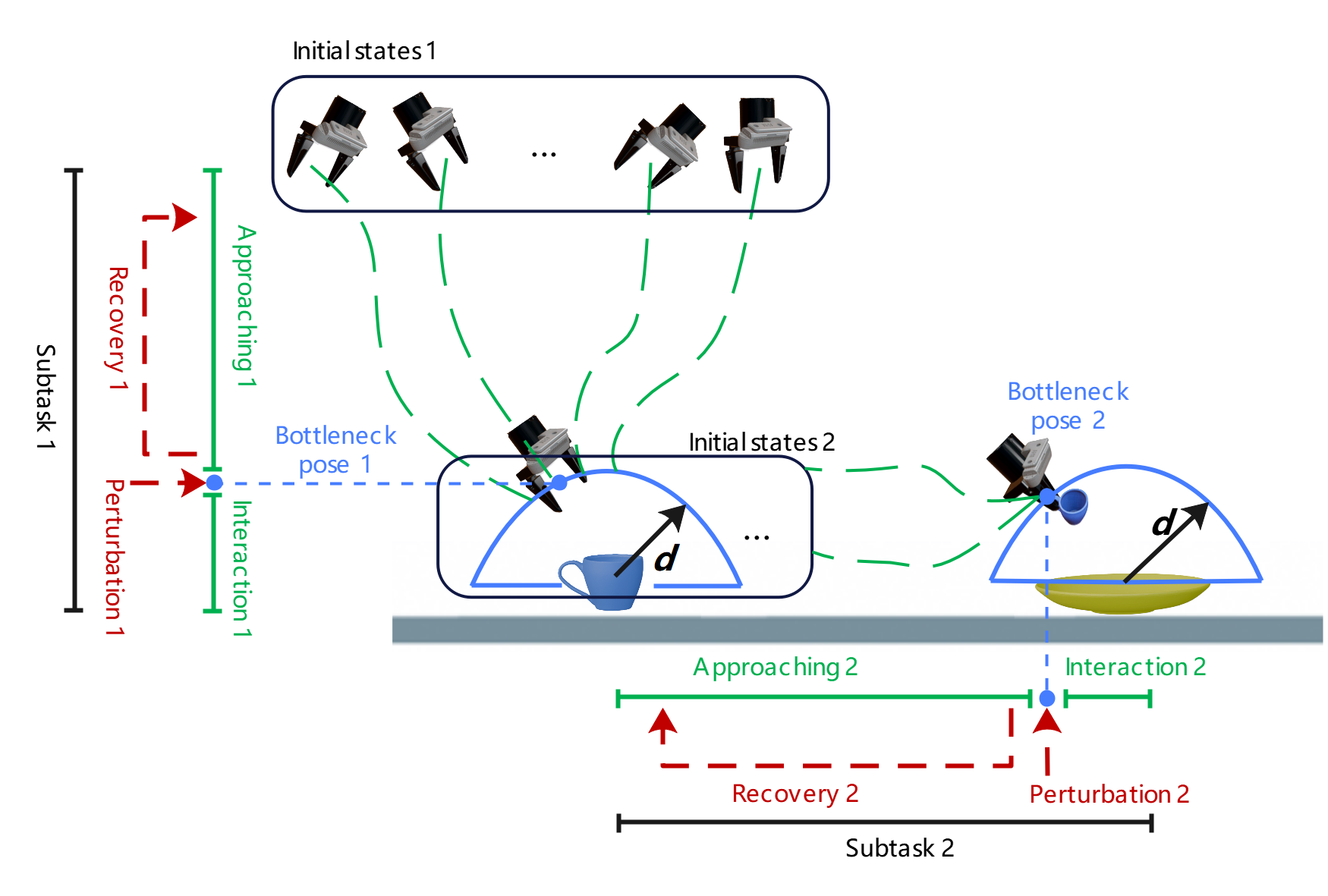}
	\caption{\textbf{Adversarial trajectory generation}. Using a cup-placement task as an  example, the manipulation is decomposed into two subtasks. Within each subtask, trajectories are synthesized piecewise, with perturbations injected at the bottleneck pose to induce robust recovery behaviors.}
	\label{fig:TrajDecomposition}
\end{figure}

Building upon this piecewise formulation, we propose a mechanism to inject abrupt perturbations into the target's state (both translation and rotation) when the gripper reaches the bottleneck pose, as illustrated in Fig.\ref{fig:TrajDecomposition}. 
This abrupt alteration of the target's state forces the motion planner to dynamically compute a recovery trajectory toward a newly updated bottleneck pose, necessitating continuous repositioning and reorientation of the end-effector.
This perturbation-recovery cycle serves a dual purpose: it significantly broadens the spatial coverage of the target's state distribution, and it exposes exteroceptive sensors (e.g., the wrist camera) to a highly diverse set of viewpoints during the recovery phase.
While these adversarial perturbations can theoretically be applied recursively, 
we empirically constrain the maximum number of interventions per trajectory to three in order to balance enhanced data diversity with motion stability and trajectory length.

\subsection{Sim-And-Real Co-Training}

In  contrast to explicit match simulation and real world, co-training treats simulation and real-world data as generic data to be sampled together, which encourages
the policies to learn invariant features across domains. 
Let $\mathcal{D}_s$ and $\mathcal{D}_r$ denote the set of simulation data and real data, respectively, where the quantity of simulation data is orders of magnitude larger than that of the real data, i.e., $|\mathcal{D}_s| \gg |\mathcal{D}_r|$. Following prior work \cite{MaddukuriCoTrainingRecipe25, WeiCotrainingDiffusionPolicy25}, we formulate the co-training problem as  minimizing the behavioral cloning loss 
\begin{equation}
    \mathcal{L}_{\mathcal{D}^{\alpha}} = \alpha  \mathcal{L}_{\mathcal{D}_s} +  (1-\alpha)  \mathcal{L}_{\mathcal{D}_r} 
\end{equation}
where $\alpha \in [0, 1]$ is the co-training ratio represents the probability of drawing a training sample from $\mathcal{D}_s$. 
In the case where $\alpha = 1$, the model is trained exclusively on simulation data, corresponding to a zero-shot deployment setting. Conversely, under the co-training paradigm, $\alpha$ is set marginally less than 1, representing a few-shot strategy where a small amount of real-world data is utilized.

\section{Experiments}
\label{sec:Experiments}

The experiments are designed to systematically validate three core hypotheses: (H1) the complete HyperSim pipeline as well as its ablated variants enable direct, zero-shot policy deployment in the real world; (H2) few-shot co-training with minimal human demonstrations synergistically improves policy performance; and (H3) the adversarial trajectory mechanism significantly enhances policy robustness against dynamic disturbances.

\subsection{Experiment Setup}

\begin{figure}[thpb]
	\centering
	\includegraphics[width=0.47 \textwidth]{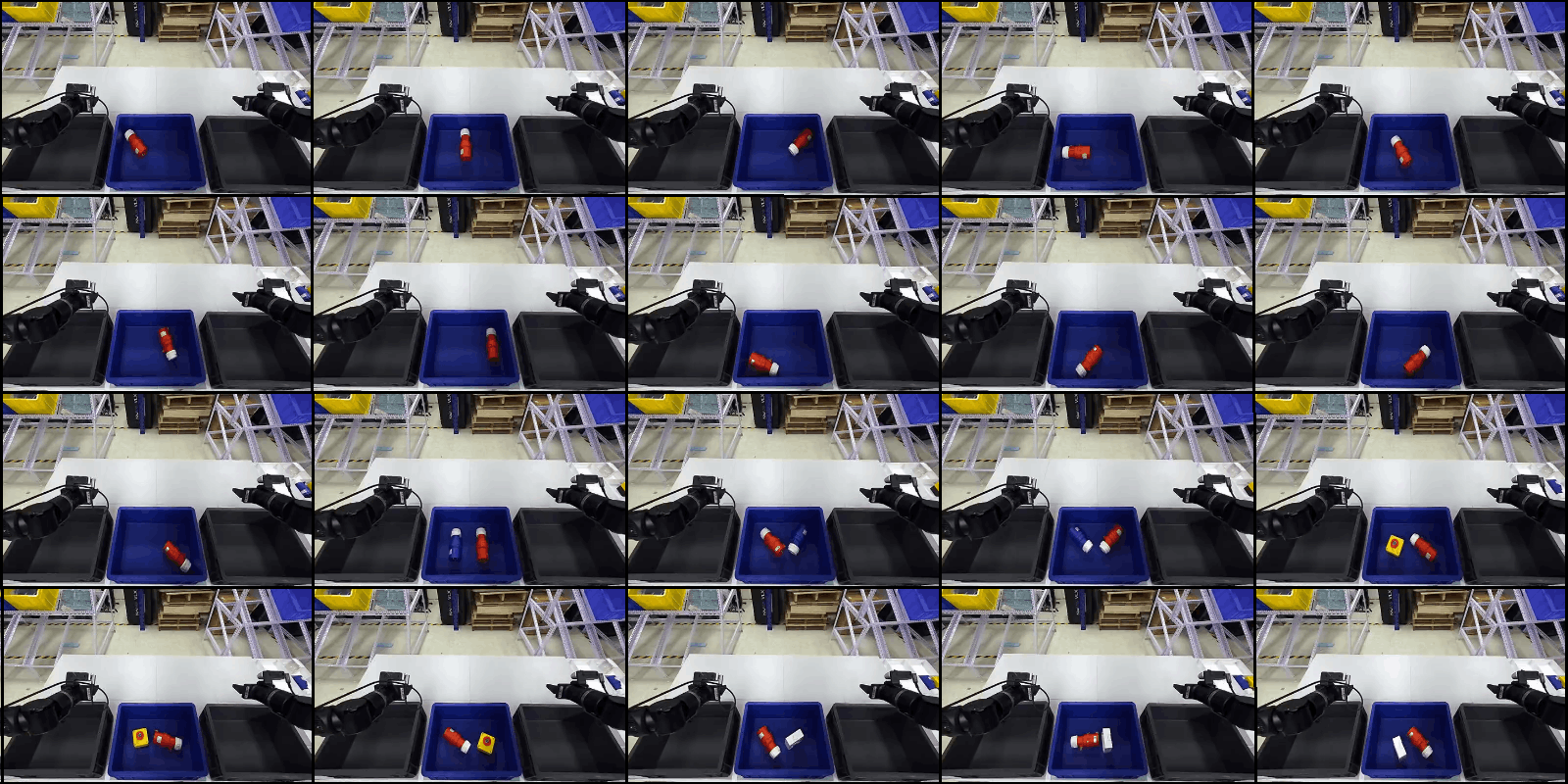}
	\caption{Initialization of 20 real-world evaluation trials with varied target poses and distractors in the central bin.}
	\label{fig:TestEnv}
\end{figure}

We conduct real-world experiments on a Galaxea R1 humanoid robot with visual observations acquired at 10 Hz from head- and wrist-mounted RGB cameras. 
The evaluation task mimics a challenging industrial sorting task that requires the robot to transfer a target object (e.g. a red plug) from a central deep bin to one of the two adjacent bins.
Note that compared to flat-surface manipulation, deep-bin picking imposes strict kinematic constraints and collision risks, particularly in corner cases where the object rests against the bin walls. 
To generate adversarial trajectories, we  perturb each component of the target 2D position by a random value uniformly drawn from $[0.02, 0.2]m$, and the target orientation is randomly sampled from the full range of $[-180, 180]^o$.
We design a fixed evaluation set of 20  trials initialized with various target poses (translation and rotation) and visual distractors to guarantee  that the evaluation is of non-trivial difficulty (Fig.\ref{fig:TestEnv}). 
Besides, this exact setup remains identical across all evaluated policies to ensure fair comparison.

\subsection{Training Datasets}
\label{sec:Dataset}

To systematically evaluate the contribution of individual modules to policy generalization as well as their compositional effect, we curate several ablated datasets alongside our holistic HyperSim pipeline. 
We utilize the simulation engine O3DE\cite{O3DE} for its native 3DGS and ROS2 support, although our framework is compatible with other popular engines such as IsaacSim and Sapien. Additionally, a minimal set of human demonstrations is collected to investigate sim-real co-training.
Overall, the training datasets include:
\begin{itemize}
    \item \textbf{BaseSim}: Standard synthetic data generated via the baseline workflow
    \item \textbf{ADSim}: Augments BaseSim with adversarial perturbation and recovery mechanisms
    \item \textbf{3DGS-ADSim}: Enhances ADSim with high-fidelity 3DGS rendering
    \item \textbf{Real$N$}: A purely real-world dataset comprising $N$ human demonstrations
    \item \textbf{Real$N$\&BaseSim} / \textbf{Real$N$\&ADSim}: Co-training mixtures combining Real${N}$ with their respective simulated counterparts
    \item \textbf{HyperSim}, or \textbf{Real${N}$\&3DGS-ADSim}: The comprehensive dataset integrating all proposed modules in our pipeline
\end{itemize}

\subsection{Evaluation Protocol}
\label{sec:EvaluationProtocol}
Robotic manipulation policies are conventionally evaluated on the basis of task success rate. However, this binary metric often fails to capture the true competence of a policy, as it treats all failure modes indiscriminately. In fact, a failure that occurs after completing  a critical step such as reaching the bottleneck pose represents a higher level of capability than a complete failure. Therefore, we introduce three fine-grained metrics to provide a more nuanced assessment:

\begin{itemize}
    \item Target Alignment Rate (TAR): The success rate of the end-effector accurately navigating from its initial state to the target's bottleneck pose.
    \item First-Attempt Success Rate ($\text{SR}_1$): The percentage of trials successfully completed in a single continuous attempt without any retries.
    \item Overall Success Rate ($\text{SR}_3$): The percentage of trials successfully completed within three attempts.
\end{itemize}

Under this protocol, we evaluated two representative policies, ACT and $\pi_0$, with a rigorous total of more than 400 real-world trials.

\section{Sim to Real Study}
\label{sec:Sim2Real}

This section details our extensive real-world empirical evaluation. We first assess the zero-shot sim-to-real transferability of our generated data.
Subsequently, we demonstrate the synergistic performance gains achieved through few-shot co-training. We conclude by subjecting the policies to dynamic environmental perturbations to rigorously validate their robustness.

\subsection{Zero-Shot Transfer}

We evaluate the zero-shot sim-to-real transfer capabilities by fine-tuning two representative policies, ACT and $\pi_0$, on datasets generated by the ablated variants of HyperSim. The quantitative results in Table \ref{tab:ZeroShotEval} yield two critical insights. 

\begin{table}[thpb]
\renewcommand{\arraystretch}{1.5}
\centering
\caption{Performance comparison of zero-shot deployments}
\begin{tabular}{c c c c c }
    \hline
    \textbf{Training Data} &\textbf{Policy}    &\textbf{TAR}     &\textbf{$\text{SR}_1$}     &\textbf{$\text{SR}_3$}    \\
    \hline
    \hline
    BaseSim                 &ACT  &10\%  & 5\%       & 5\% \\  
    ADSim                   &ACT  &45\%  & 10\%      & 15\% \\
    \textbf{3DGS-ADSim}     &ACT  &\textbf{55}\%  &\textbf{20}\%       & \textbf{25}\% \\
    \hline
    BaseSim                 &$\pi_0$  &45\%  & 45\%     & 55\% \\
    ADSim                   &$\pi_0$  &75\%  & 60\%     & 70\% \\
    \textbf{3DGS-ADSim}     &$\pi_0$  &\textbf{80}\%  & \textbf{60}\%     & \textbf{75}\% \\
    \hline
    \multicolumn{5}{l}{$^{\mathrm{*}}$3DGS-ADSim: zero-shot configuration of HyperSim}
    \end{tabular}
\label{tab:ZeroShotEval}
\end{table}

\begin{figure}[thpb]
	\centering
	\includegraphics[width=0.45\textwidth]{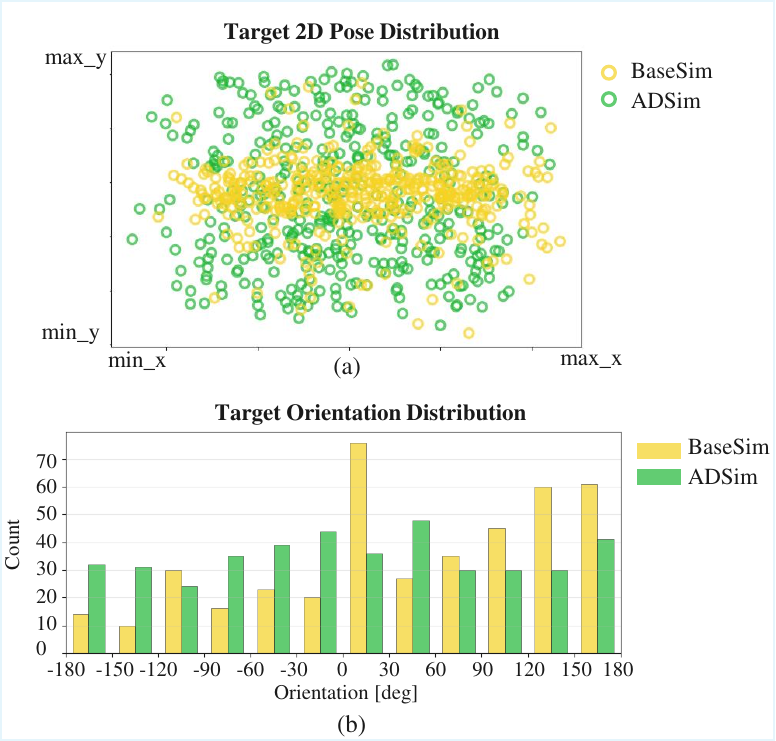}
	\caption{\textbf{Comparison of target 2D pose distribution (a) and orientation distribution (b)}.  The 2D poses in BaseSim are heavily concentrated at the workspace center, forming an elongated pattern. In contrast, ADSim expands these poses across the entire workspace to increase the spatial coverage. Similarly, the orientation distribution in BaseSim is skewed toward the $[0, 180^o]$ interval, while ADSim provides a more uniform distribution.
 }
	\label{fig:DataCoverage}
\end{figure}

\begin{figure}[thpb]
	\centering
	\includegraphics[width=0.48 \textwidth]{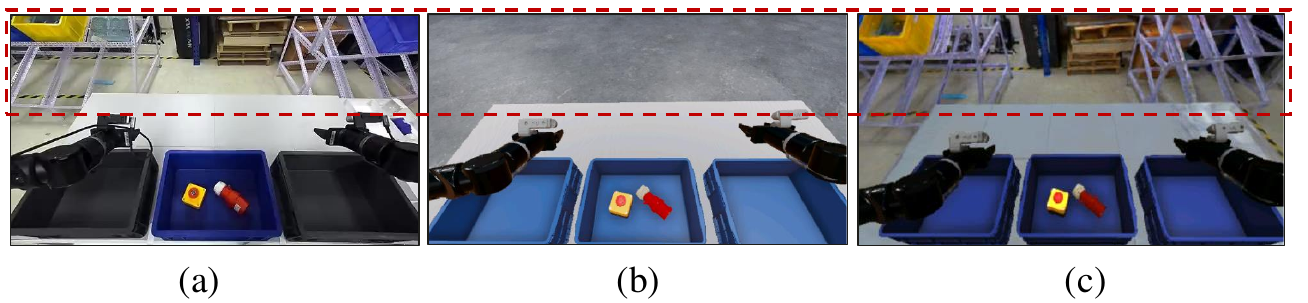}
	\caption{\textbf{Comparison of robot head-camera observations}. Compared to the real-world observation (a), BaseSim and ADSim  generate observations in a clean background (b), whereas 3DGS-ADSim (c) provides aligned observations with complex background details.}
	\label{fig:EnvComparison}
\end{figure}

\textbf{Impact of visual fidelity and adversarial diversity}. 
The progressive integration of the enhancement modules (BaseSim $\rightarrow$ ADSim $\rightarrow$ 3DGS-ADSim) consistently boosts performance. To begin with, the introduction of ADSim substantially elevates the target alignment rate (TAR). 
We posit that policies trained solely on static demonstrations tend to overfit to initial states, failing to leverage real-time visual feedback. 
In contrast, the adversarial mechanism exposes the policies to sudden target displacements at the grasping bottleneck. 
This intervention not only yields a broader and more uniform target state distribution, as shown in Fig.\ref{fig:DataCoverage}, but also forces the policies to learn closed-loop visuo-motor alignment.
Complementarily, integrating 3DGS rendering yields an additional  performance gain of up to 10\% across all metrics.
This shows that high-fidelity rendering more effectively grounds synthetic observations in physical reality, mitigating the visual domain gap (Fig.\ref{fig:EnvComparison}).
Furthermore, the drop in the first-attempt success  $\text{SR}_1$ is primarily 
caused by initial execution errors such as minor collisions with bin-walls or unstable grasps. However, with up to three retries, the policies can recover up to 15\% of performance in the overall success rate ($\text{SR}_3$).

\textbf{Synergy Between Foundation Models and Synthetic Data}.
It is interesting to note that under identical training conditions, $\pi_0$  significantly outperforms ACT by a substantial 25–55\% margin across all experiments and metrics. 
Despite the collision-prone nature of the deep-bin configuration and highly varied testing scenes, $\pi_0$ achieves an impressive $\text{SR}_3$ of 75\%.
This underscores a key takeaway: the generalization priors inherent in large-scale pre-trained models act synergistically with our high-quality synthetic data, dramatically lowering the zero-shot sim-to-real barrier.

\subsection{Few-Shot Co-Training Efficacy}

We evaluate the efficacy of few-shot co-training by integrating a small set of 35 human demonstrations to the training data. 
Results in Table. \ref{tab:FewShot} demonstrate that co-training consistently outperforms simulation-only  baselines for both ACT and $\pi_0$ policies.
We observe that the driving factors for these improvements differ based on model capacity.
For the ACT policy that exhibits limited zero-shot transferability, co-training with  real data provides essential physical grounding to effectively bridge the sim-to-real gap, especially in contact dynamics, driving a substantial increase of over 35\% in both $\text{SR}_1$ and $\text{SR}_3$ in comparison to zero-shot deployment. 
Conversely, the $\pi_0$ foundation model already exhibits strong zero-shot capabilities. In this context, co-training with real data further narrows the  sim-to-real gap. This process enhances the model's overall performance, elevating all metrics by over 15\% and bringing the  $\text{SR}_3$ to 95\%.
Moreover, co-training policies exceed the performance of their real-only counterparts (Table. \ref{tab:RealOnly}), indicating that our high-quality synthetic data effectively augment the limited number of human demonstrations for policy learning.

\begin{table}[thpb]
\renewcommand{\arraystretch}{1.5}
\centering
\caption{Performance comparison of few-shot deployments}
\begin{tabular}{c c c c c}
    \hline
    \textbf{Training Data} &\textbf{Policy}    &\textbf{TAR}     &\textbf{$\text{SR}_1$}     &\textbf{$\text{SR}_3$}    \\
    \hline
    \hline  
    ADSim           &ACT  &45\%  & 10\%        & 15\% \\
    Real35\&ADSim     &ACT  &85\%   & 65\%      & 75\% \\
   \hline
    3DGS-ADSim      &ACT  &55\%  &20\%        & 25\% \\
    \textbf{Real35\&3DGS-ADSim}     &ACT  &\textbf{85}\%   &\textbf{65}\%       & \textbf{80}\% \\
    \hline
    ADSim          &$\pi_0$  &75\%  & 60\%   & 70\% \\
    Real35\&ADSim    &$\pi_0$  &90\%  & 65\%   & 85\% \\
    \hline
    3DGS-ADSim     &$\pi_0$  &80\%  & 60\%    & 75\% \\
    \textbf{Real35\&3DGS-ADSim}    &$\pi_0$  &\textbf{95}\%  & \textbf{75}\%   & \textbf{95}\%\\
     \hline
     \multicolumn{5}{l}{$^{\mathrm{*}}$Real35\&3DGS-ADSim: few-shot configuration of HyperSim}
    \end{tabular}
\label{tab:FewShot}
\end{table}

\begin{table}[thpb]
\renewcommand{\arraystretch}{1.5}
\centering
\caption{Performance of real-only baselines}
\begin{tabular}{c c c c c}
    \hline
    \textbf{Training Data} &\textbf{Policy}    &\textbf{TAR}     &\textbf{$\text{SR}_1$}     &\textbf{$\text{SR}_3$}    \\
    \hline
    \hline
    Real35          &ACT        &70\%    &45\%      & 60\% \\
   \hline
    Real35          &$\pi_0$   &85\%    &70\%       & 70\% \\
    \hline
    \end{tabular}
\label{tab:RealOnly}
\end{table}

\subsection{Dynamic Robustness}
We design a dynamic perturbation test to assess policy robustness, an aspect often under-explored in prior work.
While scene initialization remains the same as previous experiments (Fig. 4), a human operator abruptly alters the target object's state during online policy inference. 
We record the TAR and $\text{SR}_1$ to evaluate the direct policy response against perturbations. As shown in Table.\ref{tab:Perturbation}, policies trained with adversarial trajectories (both Real35\&ADSim and Real35\&3DGS-ADSim) exhibit superior robustness with $\text{SR}_1$ being 60\% compared to 25\% achieved by the non-adversarial counterpart (Real35\&BaseSim).
This improvement stems from the adversarial mechanism, which inherently incorporates dynamic uncertainties into the training data, thereby enforcing a stronger capability to recover from  perturbations during inference.

\begin{table}[thpb]
\renewcommand{\arraystretch}{1.5}
\centering
\caption{Performance comparison against dynamic perturbations}
\begin{tabular}{c c c c c}
    \hline
    \textbf{Training Data} &\textbf{Policy}    &\textbf{TAR}     &\textbf{$\text{SR}_1$}      \\
    \hline
    \hline  
    Real35\&BaseSim     &$\pi_0$   &30\%  & 25\%  \\
    Real35\&ADSim       &$\pi_0$     &80\%   & 60\% \\
    Real35\&3DGS-ADSim   &$\pi_0$     &\textbf{80}\%   & \textbf{60}\% \\
   \hline
    \end{tabular}
\label{tab:Perturbation}
\end{table}

\section{Conclusion, Limitation, and Future work}

This paper presents HyperSim, a systematic framework designed to bridge the sim-to-real gap for vision-based robotic manipulation under zero-shot and few-shot paradigms. 
By integrating high-fidelity environment construction, adversarial trajectory generation, and sim-and-real co-training, our approach mitigates discrepancies across visual rendering, state-space distribution, and data representation. Our extensive evaluation across a suite of fine-grained metrics yields three  takeaways. First, data fidelity and coverage are pivotal to successful sim-to-real transfer. Second, large-scale pre-trained foundation models, high-quality synthetic data, and co-training exhibit profound synergies.
Third, training with adversarial data that inherently incorporates dynamic uncertainties leads to a stronger capability to recover from  perturbations during inference.

Despite these advances, we acknowledge that limitations exist in the current scope. The hardware safety constraints associated with the collision-prone deep-container manipulation, coupled with the resource-intensive nature of physical evaluation, have currently constrained our empirical validation to a specific task suite and a single humanoid embodiment. Consequently, our immediate future work will focus on scaling the HyperSim pipeline across a broader spectrum of complex manipulation tasks and diverse robotic morphologies. Ultimately, we aim to open-source the HyperSim pipeline as well as the generated dataset to further solidify the role of data synthesis as a cornerstone for generalized Embodied AI.


\section{Appendix}
\subsection{Spatial Relation Constraints For Foreground Generation}
\label{app:foreground}

We introduce 18 distinct spatial relation constraints to generate fine-grained manipulation regions in a controllable manner. The names and descriptions of these constraints are detailed in Table. \ref{tab:Constraints}.

\begin{table}[thpb]
\renewcommand{\arraystretch}{1.5}
\centering
\caption{Spatial Relation Constraints}
\begin{tabular}{l c} 
    \hline
     \textbf{Name}  &\textbf{Description}   \\
    \hline
    \hline  
    scale(OBJ, RANGE)   & scale OBJ within the RANGE  \\
    \hline
    pose2D(OBJ, RANGE)  & \makecell{randomize OBJ 2D pose within \\ the RANGE}    \\
    \hline  
    pose3D(OBJ, RANGE)  & \makecell{randomize OBJ 3D pose  within \\ the RANGE}     \\
    \hline  
    place\_on\_surface(OBJ, REF)    & \makecell{ place OBJ on top of  REFERENCE}  \\
    \hline  
   place\_left\_edge(OBJ, REF)   & \makecell{place OBJ near the left edge of \\ the REFERENCE}   \\
   \hline
    place\_right\_edge(OBJ, REF)   & \makecell{place OBJ near the right edge of \\ the REFERENCE}   \\
   \hline
    place\_top\_edge(OBJ, REF)   & \makecell{place OBJ near the top edge of \\ the REFERENCE}   \\
   \hline
    place\_bottom\_edge(OBJ, REF)   & \makecell{place OBJ near the bottom edge of \\ the REFERENCE}   \\
   \hline
    place\_to\_left(OBJ, REF)   & \makecell{place OBJ to the left of \\ the REFERENCE}   \\
   \hline
    place\_to\_right(OBJ, REF)   & \makecell{place OBJ to the right of \\ the REFERENCE}   \\
   \hline
    place\_in\_right(OBJ, REF)   & \makecell{place OBJ in front of REFERENCE}   \\
   \hline
    place\_behind(OBJ, REF)   & \makecell{place OBJ behind  REFERENCE}   \\
   \hline
    place\_front\_left(OBJ, REF)   & \makecell{place OBJ to the front-left of\\ the REFERENCE}   \\
   \hline
    place\_front\_right(OBJ, REF)   & \makecell{place OBJ to the front-right of\\ the REFERENCE}   \\
   \hline
    place\_back\_left(OBJ, REF)   & \makecell{place OBJ to the back-left of\\ the REFERENCE}   \\
   \hline
    place\_back\_right(OBJ, REF)   & \makecell{place OBJ to the back-right of\\ the REFERENCE}   \\
   \hline
    random\_placement(OBJs, REF)   & \makecell{place multiple OBJs randomly inside\\ the REFERENCE}   \\
   \hline
    no\_overlapping(OBJs, REF)   & \makecell{place multiple non-overlapped OBJs \\ inside the REFERENCE }   \\
   \hline
    with\_obstacles(OBJ, OBS, REF)   & \makecell{place OBJ with OBSTACLES \\ inside the REFERENCE}   \\
   \hline
    \end{tabular}
\label{tab:Constraints}
\end{table}

\subsection{Additional Real World Experiments}
We conducted a group of experiments comparing co-training performance with the real-only baselines. We train policies using a fixed set of 400 simulation trajectories while
varying the number of real-world demonstrations. The evaluation results are presented in Table.\ref{tab:Cotrain}. 
As expected, performance scales positively with real-world data quantity for both real-only and co-trained policies. 
More importantly, the co-trained policies consistently surpass their real-only counterpart. 
This suggests that our high-quality synthetic data act as a complementary data source,
effectively preventing the policies from overfitting to the limited real-world demonstrations.

\begin{table}[thpb]
\renewcommand{\arraystretch}{1.5}
\centering
\caption{Comparison of Real-Only and Co-Trained Policies}
\begin{tabular}{c c c c c}
    \hline
    \textbf{Training Data} &\textbf{Policy}    &\textbf{TAR}     &\textbf{$\text{SR}_1$}     &\textbf{$\text{SR}_3$}    \\
    \hline
    \hline
    Real10          &$\pi_0$  &70\%    & 35\%     & 45\% \\
    \hline  
    Real20           &$\pi_0$  &85\%  & 40\%     & 45\% \\
   \hline
    Real35          &$\pi_0$  &85\%  &70\%       & 70\% \\
    \hline
    Real10\&ADSim   &$\pi_0$  &75\%  &35\%        &55\% \\
    \hline
    Real20\&ADSim   &$\pi_0$  &85\%  &55\%        &65\% \\
    \hline
    Real35\&ADSim   &$\pi_0$  &90\%  &65\%        &85\%\\
    \hline
    Real35\&3DGS-ADSim    &$\pi_0$  &95\%  &75\%  &95\%\\
    \hline
    \end{tabular}
\label{tab:Cotrain}
\end{table}

\bibliographystyle{IEEEtran}
\bibliography{refs}

\end{document}